\documentclass{llncs}
\pdfoutput=1
\usepackage{booktabs} 
\usepackage{subdepth}
\usepackage{bm}
\usepackage{dsfont}
\usepackage{xspace}
\usepackage{algorithm}
\usepackage{amsmath}
\usepackage{amsfonts}
\usepackage{booktabs}
\usepackage{graphicx}
\usepackage{cleveref}
\usepackage{pdfpages}
\usepackage[noend]{algpseudocode}
\usepackage{todonotes}
\usepackage{pgf}
\usepackage{tikz}
\usepackage{paralist}

\usepackage{siunitx}
\usetikzlibrary{arrows,automata}

\newcommand{\Kmat}{\ensuremath{{\bm{K}}}\xspace}

\newcommand{\kvec}{\ensuremath{{\bm{k}}}\xspace}

\newcommand{\xvec}{\ensuremath{\boldsymbol{x}}\xspace}
\newcommand{\yvec}{\ensuremath{\mathbf{y}}\xspace}

\newcommand{\X}{\ensuremath{\mathcal{X}}\xspace}           

\newcommand{\R}{\ensuremath{\mathbb{R}}}                   

\newcommand{\fdist}{\ensuremath{{\text{d}}}\xspace}

\newcommand{\fkern}{\ensuremath{{\text{k}}}\xspace}

\newcommand{\fkernImpArc}{\ensuremath{{\text{k}_\text{ImpArc}}}\xspace}
\newcommand{\fdistArc}{\ensuremath{{\text{d}_\text{Arc}}}\xspace} 

\newcommand{\fdistIco}{\ensuremath{{\text{d}_\text{Ico}}}\xspace} 

\newcommand{\fdistImp}{\ensuremath{{\text{d}_\text{Imp}}}\xspace} 

\newcommand{\hats}{\ensuremath{\hat{s}}\xspace}

\newcommand{\haty}{\ensuremath{\hat{y}}\xspace}
\newcommand{\hatmu}{\ensuremath{\hat{\mu}}\xspace}

\DeclareMathOperator*{\argmin}{arg\,min}                   

\begin{document}
\title{A First Analysis of Kernels for Kriging-based Optimization in Hierarchical Search Spaces\thanks{The final authenticated version of this publication will appear in the proceedings of the 15th International Conference on
Parallel Problem Solving from Nature 2018 (PPSN XV), published in the LNCS by Springer}}

\author{Martin Zaefferer\inst{1} and Daniel Horn\inst{2}}

\institute{
TH K\"oln, Institute of Data Science, Engineering, and Analytics, \\
Steinm\"ullerallee 6, 51643~Gummersbach, Germany,
\email{martin.zaefferer@th-koeln.de}
\and
Technische Universit\"at Dortmund, Faculty of Statistics, \\
Vogelpothsweg 87,~44227~Dortmund, Germany,
\email{daniel.horn@tu-dortmund.de}
}

\maketitle

\begin{abstract}
Many real-world optimization problems require significant resources for objective function evaluations.
This is a challenge to evolutionary algorithms, as it limits the number of available evaluations.
One solution are surrogate models, which replace the expensive objective.

A particular issue in this context are hierarchical variables.
Hierarchical variables only influence the objective function if other variables satisfy some condition.
We study how this kind of hierarchical structure can be integrated into the model based optimization framework.
We discuss an existing kernel and propose alternatives.
An artificial test function is used to investigate how different kernels and assumptions affect model quality and search performance.
\end{abstract}

\section{Introduction}\label{sec:intro}
When objective function evaluations become expensive, surrogate models may be employed
to reduce the resource consumption in an optimization process.
One challenging issue in this context are \textit{conditional} or \textit{hierarchical} variables.
Hierarchical variables are only active (i.e., have an influence on the result) if other variables fulfill certain conditions.
This occurs in many algorithm tuning problems.
For instance, in machine learning algorithms, parameters of an SVM kernel are only active if that kernel is utilized.
Similarly, a variable of a variation operator in an evolutionary algorithm only has an effect if that operator is actually used.
Such parameters may also occur in engineering problems.
For instance, if a variable defining the amount of energy fed into the system exceeds a certain level, it may require an additional cooling step which itself has variables.

We require tools to model these cases efficiently.
In previous studies, three alternatives have been employed:
Firstly, the hierarchical nature of a variable could be ignored and the data handled by standard modeling methods.
This approach could be suboptimal since the available information on variable activity is not used.
Secondly, a pre-processing step could impute a constant value for the inactive variables, e.g., the mean, or some lower/upper bound~\cite{Thornton2013,Horn2016,Caceres2017a}.
We refer to this as the \textit{imputation approach}.
Thirdly, the information about hierarchical variables can be incorporated into the modeling process.
It can be be integrated into the kernel, e.g., the Arc-kernel~\cite{Hutter2013,Swersky2013}.
In other approaches, Gaussian processes are placed on the leaves of a tree structure that is assumed to represent the hierarchical dependencies of the variables~\cite{Bergstra2011,Bergstra2013,Jenatton2017}.

In this article, we focus on kernels in the context of the third case, and propose alternatives to the Arc-kernel.
We present a numerical comparison based on a simple test function to verify that the performance of these kernels meets our expectations.
We aim to answer the following research questions:

\begin{compactenum}
\item Do kernels have to incorporate knowledge about the search space hierarchy?
\item When should which kernel be used?
\item Does definiteness of the kernel play a role?
\end{compactenum}

We give a short introduction to model based optimization in Sec.~\ref{sec:SMBO} and to Kriging models in Sec.~\ref{sec:Kriging}.
Afterwards, we introduce kernels for hierarchical search spaces in Sec.~\ref{sec:hierKernels}.
We describe our experimental setup in Sec.~\ref{sec:setup} and analyze the results in Sec.~\ref{sec:results}.
A final evaluation and outlook on future work is given in Sec.~\ref{sec:out}.

\section{Surrogate Model-Based Optimization}\label{sec:SMBO}

Let $f: \X = \X_1 \times \X_2 \times \cdots \times \X_d \to \mathbb R$ be a black-box function with a $d$-dimensional input domain and a deterministic output~$y$.
Each $\X_i$ can either be numeric and bounded ($\X_i = [l_i, u_i] \subset \R$) or categorical.
We want to solve the optimization problem (OP) and find the input $\xvec^\ast = \argmin_{\xvec\in\X} f(\xvec)$.
We assume that evaluations of $f$ are expensive, which limits the number of evaluations severely.

Sequential model-based optimization (SMBO) is a state-of-the-art method for solving expensive OPs.
It is based on the Efficient Global Optimization (EGO) procedure
~\cite{Jones1998}.
First, SMBO samples and evaluates an initial set of candidate solutions.
Then, a surrogate regression model is fitted to the data.
The model is optimized with respect to an infill criterion in order to find a new, promising candidate $\xvec^*$.
The candidate $\xvec^*$ is evaluated with $f$ and added to the data set. This allows to train
an improved surrogate model.
The procedure iterates until a stopping criterion is reached, e.g., a budget on the number of function evaluations.
A detailed introduction is given by Bischl et al. in \cite{bischl2017}.

Four components of the SMBO procedure have to be specified: the generation of the initial candidate set, the surrogate model, the infill criterion and the optimizer of the infill criterion.
We use Latin Hypercube Sampling (LHS), Kriging models, the expected improvement criterion and Differential Evolution (DE)~\cite{Storn1997}. 
Our methods can be easily extended to other SMBO variants that employ kernel-based models.

\section{Kriging}\label{sec:Kriging}
Frequently, SMBO employs Kriging models, which interpret observations as realizations of a Gaussian process.
Forrester et al.~\cite{Forrester2008a} give a detailed description.
In its core, Kriging models the correlation between observations, e.g., with an exponential correlation function $\fkern(\xvec,\xvec')=\exp(-\theta \cdot \fdist(\xvec,\xvec'))$.
Here, $\xvec$  and $\xvec'$ are samples, $\theta$ is a kernel parameter and $\fdist(\xvec,\xvec')$ is a distance  function, e.g., the Euclidean distance if $\xvec$ is real valued.
The correlation matrix \Kmat collects all pairwise correlations.
Usually, correlation functions should be positive semi-definite (PSD), i.e., all eigenvalues of \Kmat are non-negative.
The Kriging predictor is:
\begin{align*}
\haty(\xvec)=\hat{\mu}+\kvec^T \Kmat^{-1} (\yvec-\bm{1}\hat{\mu}),
\end{align*}
where \yvec are the training observations, \hatmu represents the process mean, $\bm{1}$ is a vector of ones and \kvec is the column vector of correlations between the set of training samples
$\bm{X}$ and the new sample $\xvec$.
All parameters are usually determined by Maximum Likelihood Estimation (MLE).
Kriging is a popular choice in SMBO algorithms, as it provides an estimate of the prediction uncertainty:%
\begin{align*}
\hats^2(\xvec)= \hat{\sigma}^2  (1- \kvec^T \Kmat^{-1}\kvec ),
\end{align*}
where the process variance $\hat{\sigma}$ is also determined by MLE.
The estimate $\hats^2(\xvec)$ can be used to balance exploration and 
exploitation by computing the Expected Improvement (EI) of candidate solutions~\cite{Mockus1978}.
The EI is a frequently employed infill criterion, e.g., in EGO~\cite{Jones1998}.

Kriging also allows to deal with noisy data, using the so called nugget effect.
The nugget adds a small constant $\eta > 0$ to the diagonal of \Kmat.
Thus, the otherwise interpolating Kriging model is able to regress the data, introducing additional smoothness into the predicted value.
The nugget effect may also help to increase the numerical stability.
A re-interpolation approach can be used to avoid that the nugget effect deteriorates the uncertainty estimate~\cite{Forrester2008a}.

\section{Kernels for Hierarchical Search Spaces}\label{sec:hierKernels}
Hierarchical variables can be defined as variables that are only \textit{active} if other variables fulfill a condition.
An \textit{active} variable has an impact on the objective function value.
We use the notation of Hutter and Osborne~\cite{Hutter2013}: a function $\delta_i(\xvec)$ determines whether the $i$-th variable of $\xvec$ is active ($\delta_i(\xvec)=\text{true}$) or not (false).
In the following, only the per-variable distance $\fdist_i(x_i,x_i')$ will be introduced for each kernel. 
The combined kernel structure is identical for all cases unless stated otherwise, i.e., $\fkern(\xvec,\xvec')=\exp(-\sum_{i=1}^d \fdist_i(x_i,x_i') )$. We describe an existing kernel (Arc) and propose four alternatives (Ico, IcoCorrected, Imp, ImpArc).

\subsection{The Arc-kernel}
The Arc-kernel proposed by Hutter and Osborne~\cite{Hutter2013} 
is specifically developed to handle hierarchical structures.
It is based on three assumptions.
First, if a hierarchical variable is inactive in two configurations $\xvec$ and $\xvec'$, then the distance in that dimension should be zero.
Second, if it is active in both configurations, the distance depends on the respective variable values.
Third, if the variable is only active in one configuration, the distance should be a constant, because
no information is available to compare an inactive with an active variable.

An embedding is required to encode these assumptions in valid distance measures that yield a PSD kernel.
It is for continuous variables~\cite{Hutter2013}:
\begin{align}\label{eq:arc1}
\fdistArc_i(x_i,x_i')  &= \left\{
  \begin{array}{l l}
   0, & \text{if } \delta_i(\xvec) = \delta_i(\xvec') =\text{false}\\
   \theta_i, &  \text{if } \delta_i(\xvec) \neq \delta_i(\xvec')\\
   \theta_i \sqrt{2-2\cos(\pi \rho_i \frac{x_i-x_i'}{u_i - l_i}}),  &  \text{if } \delta_i(\xvec) = \delta_i(\xvec') =\text{true}\\
  \end{array} \right.
\end{align}
The kernel variables $\theta_i \in \mathbb{R}^+$ and $\rho_i \in [0,1]$ are determined by MLE.
A respective measure for categorical variables can be found in~\cite{Hutter2013}.
We follow up on~\cite{Swersky2013} and skip the notion of putting further restrictions on $\theta_i$ to encode lower importance of lower hierarchical levels as proposed in~\cite{Hutter2013}.
Moreover, we use the square of the distance in the embedded space (i.e., removing the square root in Eq.~\eqref{eq:arc1}),
since we also use squared deviations in all other distances.

\subsection{Indefinite Conditional Kernel}
We propose a simplified alternative to the Arc-kernel:
\begin{align*}
\fdistIco_i(x_i,x_i')  &= \left\{
  \begin{array}{l l}
   0, &  \text{if } \delta_i(\xvec) = \delta_i(\xvec') =\text{false}\\
   \rho_i, &  \text{if } \delta_i(\xvec) \neq \delta_i(\xvec')\\
   \theta_i \fdist_i(x_i,x_i'), &  \text{if } \delta_i(\xvec) = \delta_i(\xvec') =\text{true}\\
  \end{array} \right.
\end{align*}
Here, $\fdist_i(x_i,x_i')$  is an appropriate default distance (numerical: square deviation $(x_i-x_i')^2$, categorical: Hamming distance).
The distance parameter $\rho_i \in \mathbb{R}^+$ is determined by MLE.
The kernel follows the same intuitive assumptions as \fdistArc, but it does not use the complicated cylindrical embedding.
This may lead to indefinite kernel matrices for some data sets or choices of parameters.
Due to this, it will be denoted as the indefinite conditional kernel, or Ico-kernel.

As a variant of the Ico-kernel, the IcoCorrected (IcoCor) kernel is the same kernel subject to a correction via a spectrum-flip.
This transformation of the eigenspectrum generates PSD kernel matrices from indefinite kernels, cf.~\cite{Zaefferer2016b}.
Note, that the nugget effect may also correct issues with definiteness if $\eta$ is large
enough. Thus, even the uncorrected Ico-kernel can produce a valid model.

\subsection{Imputation Kernel} 
Alternatively, we propose a simple PSD kernel.
It is based on a different assumption: If the hierarchical variable is only active in one of two configurations ($\delta_i(\xvec) \neq \delta_i(\xvec')$), their distance in that dimension is \textit{not} assumed to be constant.
Rather, it is assumed that the value of the active configuration does influence the dissimilarity.
This is achieved by introducing a kernel parameter against which the respective active value is compared.
Thus,
\begin{align*}
\fdistImp_i(x_i,x_i')  &= \left\{
  \begin{array}{l l}
    0, & \text{if } \delta_i(\xvec) = \delta_i(\xvec') =\text{false}\\
   \theta_i \fdist_i(x_i',\rho_i), &  \text{if } \delta_i(\xvec) = \text{false} \neq \delta_i(\xvec')\\
   \theta_i \fdist_i(x_i,\rho_i), &  \text{if } \delta_i(\xvec) = \text{true} \neq \delta_i(\xvec')\\
   \theta_i \fdist_i(x_i,x_i'), &  \text{if }\delta_i(\xvec) = \delta_i(\xvec') =\text{true}\\
  \end{array} \right.
\end{align*}
where $\fdist_i$ is again the appropriate default distance (square deviation, Hamming) and $\rho_i$ is of the same data type as $x_i$.
For real $x_i$, the bounds of $x_i$ and $\rho_i$ can differ. We use  $\rho_i \in [l_i-a, u_i+a] \subset \R$ with $a=2*(u_i-l_i)$. Larger bounds may be necessary, depending on the problem.
Similarly, if $x_i$ is categorical  $\rho_i$ can have one more level (category) than $x_i$, to emulate the case where none of the other levels is a good replacement.
An exponential kernel based on $\fdistImp_i(x_i,x_i')$ can be proven to be PSD.
Using proposition 2 in~\cite{Hutter2013}, we only need to show that there exists a mapping function $f_i(x_i)$ that maps to a space in which a valid distance can be used, i.e., $\fdistImp_i(x_i,x_i') = \fdist_i(f_i(x_i),f_i(x_i'))$.
For \fdistImp, the mapping function is
\begin{align*}
f_i(x_i) &= \left\{
  \begin{array}{l l}
   x_i &\text{if } \delta_i(\xvec)\\
   \rho_i & \text{otherwise.}
  \end{array} \right.
\end{align*}
Hence, the resulting kernel based on $\fdist_i(f_i(x_i),f_i(x_i'))$ is PSD.

Clearly, this kernel has relations to the imputation approach mentioned in Sec. \ref{sec:intro}.
Essentially, inactive values are replaced by an imputed value $\rho_i$.
Instead of choosing that value a-priori, it is defined as a parameter and determined by MLE.
Hence, it will be denoted as the imputation kernel or Imp-kernel.
One drawback of this kernel is, that if $x_i$ is categorical, $\rho_i$ is also categorical.
This may complicate the MLE procedure.
Also, the assumption that some value can be imputed is less conservative than the assumptions of the Arc-kernel.

\subsection{The Imputation-Arc Kernel}
When it is unclear whether the Arc- or Imp-kernel is more appropriate, we suggest a linear combination denoted as the ImpArc-kernel, 
\begin{align*}
\fkernImpArc(\xvec,\xvec')=\exp(-\sum_{i=1}^d\beta_{1,i} \fdistArc_i(x_i,x_i') + \beta_{2,i} \fdistImp_i(x_i,x_i') ) ,  
\end{align*}
with weights $\beta_{k,i} \in \mathbb{R}^+$ determined by MLE.
Other combinations (e.g., Ico-Imp, Imp-Arc-Ico) are possible. 
We only test the ImpArc combination, because the Ico- and Arc-kernel express very similar information. 
Also, a three-way combination would require to learn an additional weight $\beta_{3,i}$.

\section{Experimental Setup}\label{sec:setup}

\begin{figure}[!b]
\centering
\includegraphics[width=\textwidth]{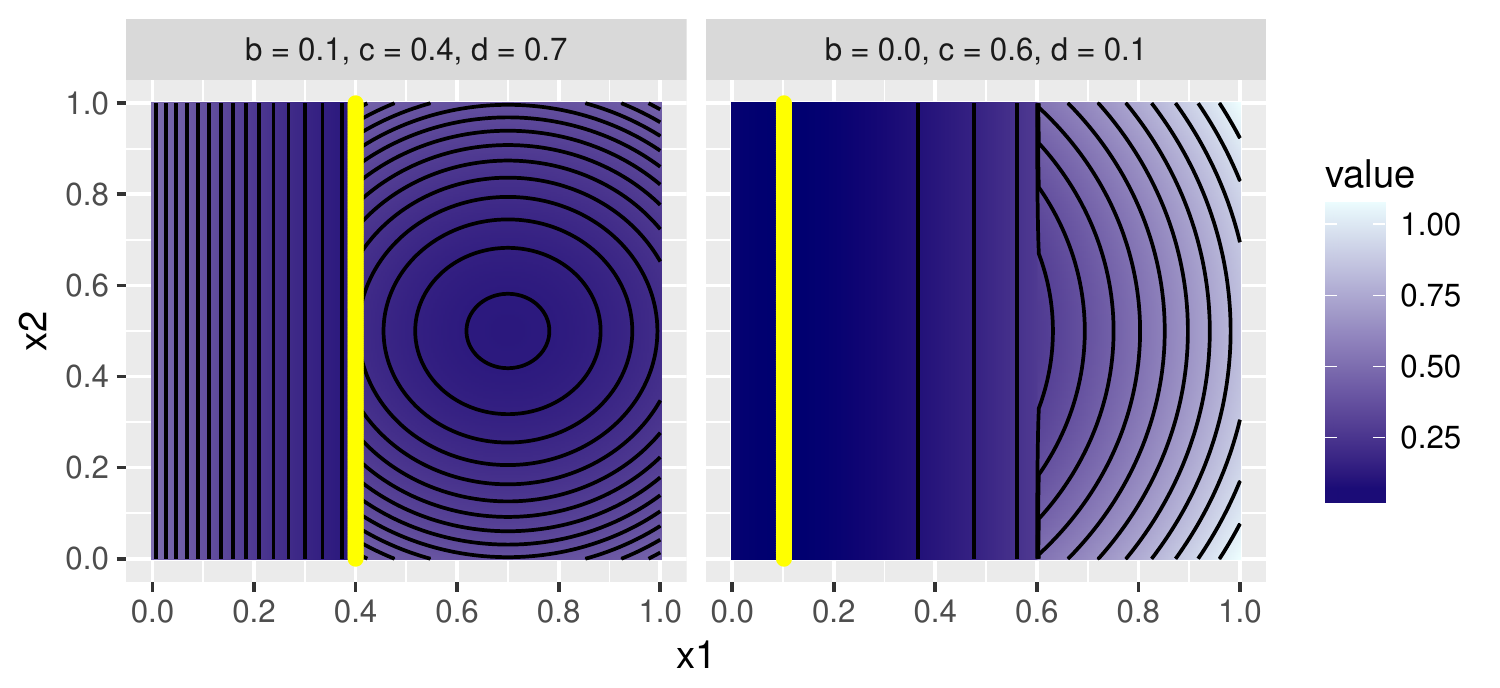}
\caption{Visualization of the test function, the optimum is marked in yellow.}
\label{fig:testfun}
\end{figure}
While synthetic, tree-based test functions for hierarchical search spaces have 
been proposed by Jenatton et al.~\cite{Jenatton2017},
they are not able to respect the different definitions and assumptions of our kernels. 
Hence, we suggest a simple two-dimensional quadratic function
\begin{align*}
f(\xvec) =(x_1 - d)^2  + \left\{
  \begin{array}{l l}
   0 & \text{if} ~~~ x_1 \leq c\\
   (x_2 - 0.5)^2 + b &  \text{else}\\
  \end{array}\right. .
\end{align*}
The function's behavior (see Fig. \ref{fig:testfun}) is defined by the constants $b, c$ and $d$. 
The constant $b$ controls whether the Imp-kernel is a good match, $c$ controls the size of the active region and $d$ controls the location of the optimum.
The function is influenced by the hierarchical variable $x_2$ only if $x_1 > c$ and does have a discontinuity at $x_1 = c$.
For $b=0$, the function is continuous at $x_2 = 0.5$. Hence, the if-else term of $f(\xvec)$ yields identical results
 if $x_2=0.5$ and if $\delta_2(\xvec)=false$.
In this case, the assumption of the Imp-kernel is fulfilled, i.e., 
the kernel definition matches the problem structure.
The Imp-kernel should learn to impute $\rho=0.5$.

We identified five situations with different expected performances.
\begin{compactenum}[A)]
\item
$d<c$ (the optimum is in the \textit{inactive} region of $x_2$ at $x_1 = d, x_2 \in \R$) and $b=0$ (imputation potentially \textit{profitable}). The function is \textit{unimodal}.
\item
$d<c$ (the optimum is in the \textit{inactive} region at $x_1 = d, x_2 \in \R$) but $b>0$ (imputation potentially \textit{unprofitable}). The function is \textit{unimodal}.
\item
$d > c$ (the optimum is in the \textit{active} region at $x_1 = d, x_2 = 0.5$) and $b=0$ (imputation potentially \textit{profitable}). The function is \textit{bimodal}.
\item
$d > c$ (the optimum is in the \textit{active} region at $x_1 = d, x_2 = 0.5$) and $b=0.1$ (imputation potentially \textit{unprofitable}) and $b< (c-d)^2$. 
The function is  \textit{bimodal}.
The discontinuity at $c$ is not as important, since the optimum is remote from it.
\item
$d > c$ (the optimum is in the \textit{active} region at $x_1 = c, x_2 \in \R$) and $b=0.1$ (imputation potentially \textit{unprofitable}) and $b> (c-d)^2$. 
The function is  \textit{bimodal}.
The discontinuity at $c$ has to be approximated well, since the optimum is at $x_1 = c$.
\end{compactenum}
Covering all of these five situations, we tested all combinations of the values \mbox{$b=\{0, 0.1\}$}, $c=\{0.2, 0.4, 0.6, 0.8\}$, and $d=\{0.1, 0.3, 0.5, 0.7, 0.9\}$.


To estimate model quality, we measured the model's Root Mean Squared Error (RMSE).
The models were trained with 10, the error was estimated on 1\,000 uniform random samples.
The Kriging model was trained with the \texttt{CEGO} package in R~\cite{CEGOv2.2.0,Zaefferer2014b}.
It was configured to use the nugget effect and re-interpolation.
The Dividing Rectangles algorithm~\cite{Jones1993} was chosen to optimize the model parameters 
via 200 likelihood evaluations.
We used all kernels from Sec.~\ref{sec:hierKernels} and a standard exponential kernel with square deviation in each dimension (which does not incorporate hierarchical information), denoted as the Stan-kernel.

The same type of model was used in the SMBO algorithm
from the \texttt{CEGO} package. The search was limited to 10 evaluations of $f(\xvec)$, 
due its low difficulty, low dimensionality and assumed cost.
The search was initialized with three uniform random samples. 
Based on the model, the EI criterion was optimized by DE~\cite{Storn1997}. 
We used the \texttt{DEoptim} package~\cite{Mull11} with $10\,000$ EI evaluations per iteration and used default parameters otherwise.
Each experiment was repeated 100 times, with 100 unique random seeds (one per replication).
We recorded the difference between the best found and the optimal function value (\emph{suboptimality}) for each replication.

\section{Results}\label{sec:results}
\begin{figure}[bt]
\centering
\includegraphics[width=0.9\textwidth]{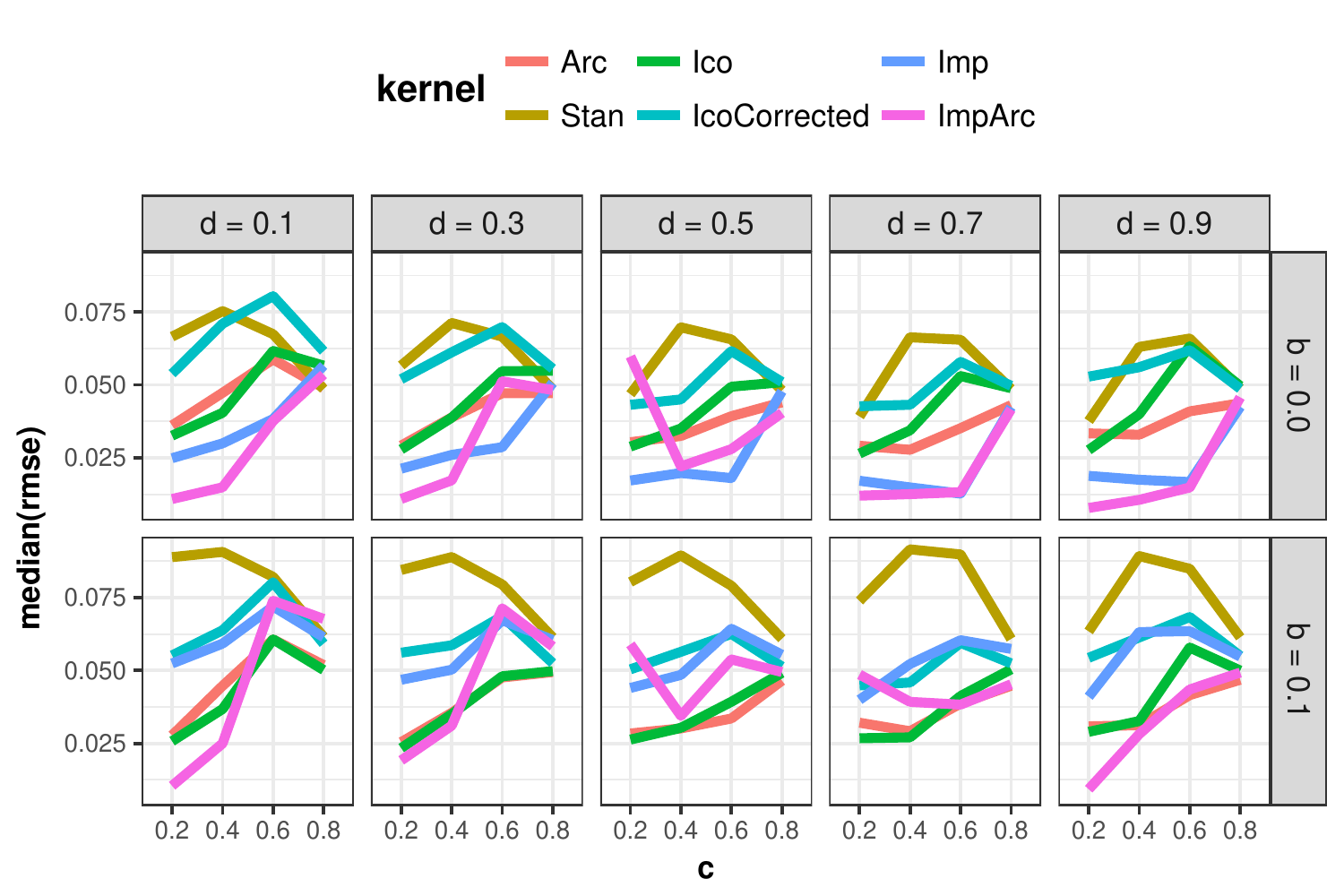}
\caption{Median RMSE values over 100 replications for all configurations and kernels.}
\label{fig:mod_qual_plot}
\end{figure}

First, we analyze the model quality produced by the different kernels.
Fig. \ref{fig:mod_qual_plot} shows the median RMSE value for all parameter constellations and kernels.
Clearly, the fit of the Stan-kernel is inferior to most specialized hierarchical kernels for almost all parameter constellations, especially  if $b = 0.1$.

If $b = 0$, the assumption of the Imp-kernel is fulfilled.
Hence, both the Imp- and the ImpArc-kernel produce a better fit than most other kernels.
However, for $b = 0.1$, the Imp-kernel mostly has the second or third worst performance. Only the Stan-kernel and sometimes the IcoCorrected-kernel perform worse.
The Arc- and the Ico-kernel achieve very similar performances in most cases, with near-to-best performance if $b = 0.1$.
The ImpArc-kernel, combining the advantages of the Arc- and the Imp-kernel, has a good, sometimes best fit in all situation, for both $b \in \{0, 0.1\}$.
Contrarily, the IcoCorrected-kernel has a rather poor fit in several cases, sometimes even worse than the Stan-kernel.
Overall, differences between kernels tend to disappear for large values of $c$, which is to be expected due to the reduced influence of the hierarchical variable $x_2$.

\begin{figure}[tb]
\centering
\includegraphics[width=\textwidth]{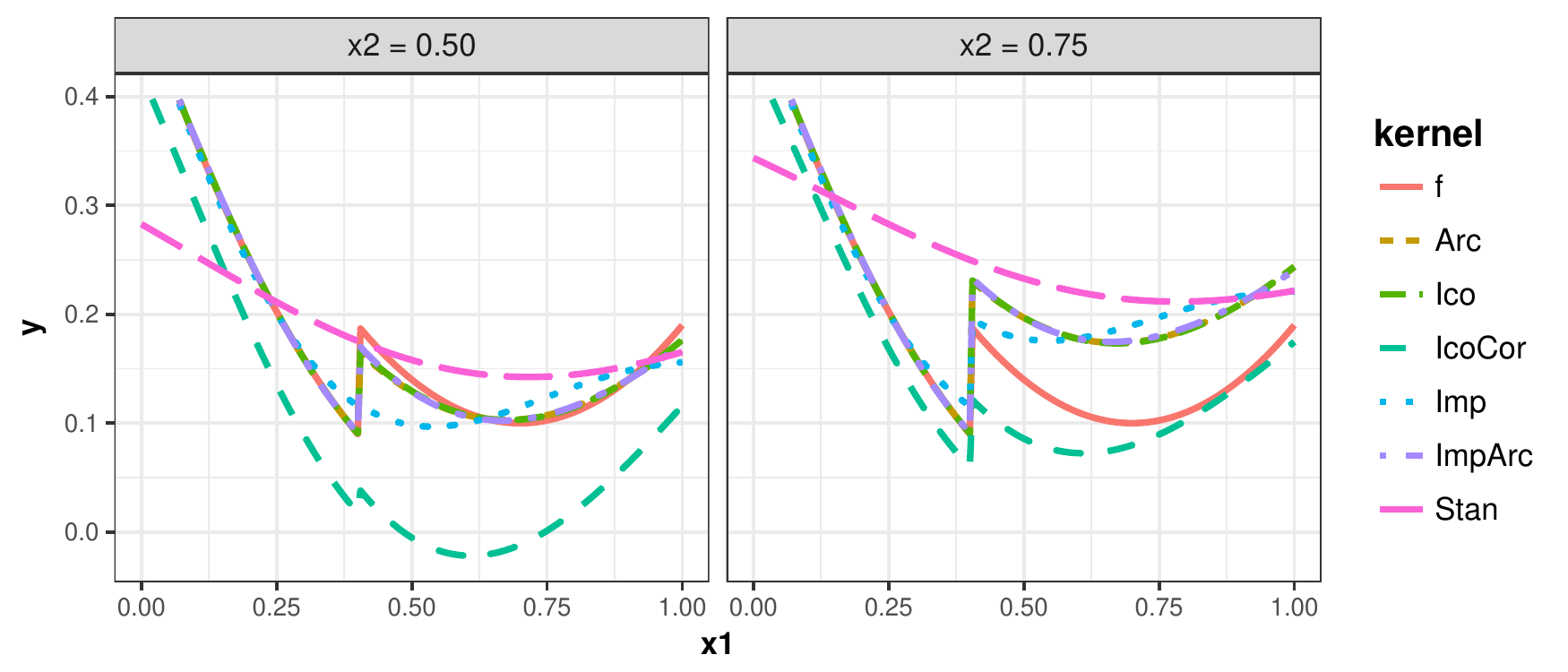}
\caption{Example fits of the kernels. Two slice planes are shown for $x_2 \in \{0.5, 0.75\}$.}
\label{fig:line_plots}
\end{figure}

To get a better understanding of the kernels, we visualize an example for Situation E with $(b, c, d) = (0.1, 0.4, 0.7)$.
Fig.~\ref{fig:line_plots} shows line plots for the test function as well as fitted models for all six kernels, trained with ten uniform random samples.
Here, the global optimum is at $x_1 = c = 0.4$, i.e., at the jump discontinuity.
The function value of the global optimum ($0.09$) is only slightly better than the value of the local optimum ($0.1$) at $(x_1 = 0.7, x_2=0.5)$.
Hence, to find the global optimum, it is important to model the discontinuity well.

The Stan-kernel is not able to model the discontinuity and therefore tries to fit a smooth curve to the function.
Hence, the Stan-kernel approximates the optimum poorly.
For $x_1 = 0.5$ the model of the Imp-kernel shares the poor performance of the Stan kernel: It is not able to fit the discontinuity.
Still, the fit is much closer to the true objective function.
For $x_1 = 0.75$ the Imp-kernel is able to fit the discontinuity, but the fit is inferior to the Arc-, Ico- and ImpArc-kernel.
All of them reproduce the discontinuity quite well.
However, their approximation of the function for $x_1 > c, x_2 = 0.75$ has a strong offset.
While this is not a perfect fit, it will not necessarily deteriorate optimization performance.
The model based on the IcoCorrected-kernel is able to reproduce the discontinuity, but the jump is not large enough to identify the optimum at $x_1 = 0.4$.

Next, we analyze the optimization performance.
Due to space restrictions, we present statistical test results that summarize the experimental data.
Following Dem\v{s}ar~\cite{Demsar2006}, we apply Friedman and corresponding post-hoc Nemenyi tests in order to find significant differences between the kernels, using the function parameters $b, c$ and $d$ as blocking variables for the tests.
We extend Dem\v{s}ar's approach, since we do not apply our tests to the median suboptimalities.
Instead, we use the replication identifier as an additional blocking variable.
This accounts for the effect of the initial design.
We visualize the test results using ordered graphs that present a rough order on the kernels.

We start by investigating the combined results of all optimization experiments.
With a p-value that is numerically approximating zero ($<10^{-16}$), the Friedman-test indicates that there are significant differences between the different kernels. 
Note, if differences are present p-values tend to be small due to the large number of experiment replications, and differentiating between significant and relevant differences is an open issue in the analyses of computer experiments.

\begin{figure}[!b]
   \fbox{
   \begin{minipage}{.46\linewidth}
\begin{tikzpicture}[->,>=stealth',shorten >=1pt,auto,node distance=2.8cm,
                    semithick]
  \tikzstyle{every state}=[rectangle, align = center, font=\scriptsize]
  \tikzstyle{every path}=[align = center, font=\scriptsize]

  \node[state] (Imp)    at(0, 0.5)     {Imp    \\ 3.20};
  \node[state] (ImpArc) at(0, 1.5)     {ImpArc \\ 3.23};
  \node[state] (Ico)    at(2, 1)     {Ico    \\ 3.40};
  \node[state] (Arc)    at(2, 0)     {Arc    \\ 3.41};
  \node[state] (IcoCor) at(2, 2) {IcoCor \\ 3.41};
  \node[state] (Stan)   at(4, 1) {Stan   \\ 4.35};

  \path (Imp)    edge  [densely dotted] (Ico)
                 edge  [densely dotted] (IcoCor)
                 edge  [densely dotted] (Arc)
        (ImpArc) edge  [densely dotted] (Ico)
                 edge  [densely dotted] (IcoCor)
                 edge  [densely dotted] (Arc)
        (IcoCor) edge                   (Stan)
        (Ico)    edge                   (Stan)
        (Arc)    edge                   (Stan);
\end{tikzpicture}
  \end{minipage}%
  }
  \hfill%
  \begin{minipage}{.46\linewidth}
\begin{tikzpicture}[->,>=stealth',shorten >=1pt,auto,node distance=2.8cm,
                    semithick]
  \tikzstyle{every state}=[draw=none,fill=none, align = center]
  \tikzstyle{every path}=[align = center]

  \node[draw=none,fill=none, font = \large] at (0.5, 2.75) {Legend};

  \node[] at (0, 2) (d01) {};
  \node[] at (0, 1.5) (d11) {};
  \node[] at (0, 1) (d21) {};
  \node[] at (0, 0.5) (d31) {};
  \node[right = 1cm of d01] (d02) {};
  \node[right = 1cm of d11] (d12) {};
  \node[right = 1cm of d21] (d22) {};
  \node[right = 1cm of d31] (d32) {};
  \node[draw=none,fill=none, right = 0cm of d02] {significant to niveau $10^{-12}$};
  \node[draw=none,fill=none, right = 0cm of d12] {significant to niveau $10^{-6}$};
  \node[draw=none,fill=none, right = 0cm of d22] {significant to niveau $0.01$};
  \node[draw=none,fill=none, right = 0cm of d32] {significant to niveau $0.1$};

  \path (d01) edge                    (d02)
        (d11) edge   [dashed]         (d12)
        (d21) edge   [densely dotted] (d22)
        (d31) edge   [loosely dotted] (d32);
\end{tikzpicture}

  \end{minipage}%
\caption{Ordering of the six kernels with respect to their mean ranks (printed below each kernel) over all test instances.  A path (possibly using multiple edges) between two kernels denotes a significant difference of the post-hoc Nemenyi test, the directions of the arrows follows the ordering of the mean ranks.}

\label{fig:graphs_all}
\end{figure}
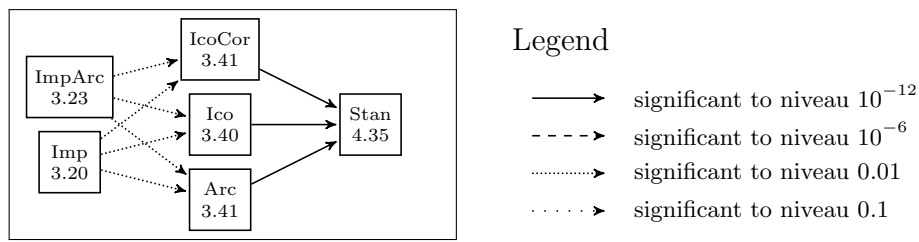

Fig. \ref{fig:graphs_all} shows the results of the corresponding Nemenyi-test, including a graph representation of the test results as well as mean ranks for each kernel.
As expected, the Stan-kernel is clearly outperformed by all other kernels.
For the other kernels, we can identify two groups: The Imp- and the ImpArc-kernel seem to perform slightly better than the rest.
Within each group, there are no significant differences between the kernels, while tests between kernel from groups are significant.
It is questionable how reliable this result is. 
We expect diverse behavior of the kernels in the five situation and the overall performance is of course influenced by the selection of the specific test instances.
Hence, we will now examine individual tests for situations A to E.

\begin{figure}[!b]
 \fbox{
 \begin{minipage}[t]{.46\linewidth}
\begin{tikzpicture}[->,>=stealth',shorten >=1pt,auto,node distance=2.8cm,
                    semithick]
  \tikzstyle{every state}=[rectangle, align = center, font=\scriptsize]
  \tikzstyle{every path}=[align = center, font=\scriptsize]

  \node[font = \large\bf] at (-0.25, 2.2) {A};

  \node[state] (Imp)    at(0, 1)     {Imp    \\ 3.01};
  \node[state] (ImpArc) at(2, 2)     {ImpArc \\ 3.29};
  \node[state] (Ico)    at(2, 1)     {Ico    \\ 3.35};
  \node[state] (Arc)    at(2, 0)     {Arc    \\ 3.41};
  \node[state] (IcoCor) at(4, 1.625) {IcoCor \\ 3.90};
  \node[state] (Stan)   at(4, 0.375) {Stan   \\ 4.03};

  \path (Imp)    edge  [loosely dotted] (ImpArc)
                 edge  [densely dotted] (Ico)
                 edge  [dashed]         (Arc)
        (ImpArc) edge                   (IcoCor)
                 edge                   (Stan)
        (Ico)    edge  [dashed]         (IcoCor)
                 edge                   (Stan)
        (Arc)    edge  [dashed]         (IcoCor)
                 edge                   (Stan);
\end{tikzpicture}
  \end{minipage}%
  }
  \hfill%
  \fbox{
  \begin{minipage}[t]{.46\linewidth}
\begin{tikzpicture}[->,>=stealth',shorten >=1pt,auto,node distance=2.8cm,
                    semithick]
  \tikzstyle{every state}=[rectangle, align = center, font=\scriptsize]
  \tikzstyle{every path}=[align = center, font=\scriptsize]

  \node[font = \large\bf] at (4.25, 2.2) {B};

  \node[state] (Imp)    at(2, 1) {Imp    \\ 3.52};
  \node[state] (ImpArc) at(2, 2) {ImpArc \\ 3.41};
  \node[state] (Ico)    at(0, 1) {Ico    \\ 2.89};
  \node[state] (Arc)    at(0, 0) {Arc    \\ 3.02};
  \node[state] (IcoCor) at(2, 0) {IcoCor \\ 3.55};
  \node[state] (Stan)   at(4, 1) {Stan   \\ 4.62};

  \path (Ico)    edge  [dashed]         (ImpArc)
                 edge                   (Imp)
                 edge                   (IcoCor)
        (Arc)    edge  [densely dotted] (ImpArc)
                 edge  [dashed]         (Imp)
                 edge  [dashed]         (IcoCor)
        (ImpArc) edge                   (Stan)
        (Imp)    edge                   (Stan)
        (IcoCor) edge                   (Stan);
\end{tikzpicture}
  \end{minipage}%
  }

\vspace{0.2cm}

 \fbox{
 \begin{minipage}{.46\linewidth}
\begin{tikzpicture}[->,>=stealth',shorten >=1pt,auto,node distance=2.8cm,
                    semithick]
  \tikzstyle{every state}=[rectangle, align = center, font=\scriptsize]
  \tikzstyle{every path}=[align = center, font=\scriptsize]

  \node[font = \large\bf] at (-0.25, 3.2) {C};

  \node[state] (Imp)    at(0, 2)   {Imp    \\ 2.28};
  \node[state] (ImpArc) at(1.5, 2) {ImpArc \\ 3.01};
  \node[state] (Ico)    at(4.5, 1) {Ico    \\ 4.18};
  \node[state] (Arc)    at(3, 1)   {Arc    \\ 3.77};
  \node[state] (IcoCor) at(3, 3)   {IcoCor \\ 3.64};
  \node[state] (Stan)   at(4.5, 3) {Stan   \\ 4.13};

  \path (Imp)    edge                  (ImpArc)
        (ImpArc) edge                  (IcoCor)
                 edge                  (Arc)
        (IcoCor) edge [dashed]         (Ico)
                 edge [dashed]         (Stan)
        (Arc)    edge [densely dotted] (Ico)
                 edge [densely dotted] (Stan);
\end{tikzpicture}
  \end{minipage}%
  }
  \hfill%
   \fbox{
  \begin{minipage}{.46\linewidth}
\begin{tikzpicture}[->,>=stealth',shorten >=1pt,auto,node distance=2.8cm,
                    semithick]
  \tikzstyle{every state}=[rectangle, align = center, font=\scriptsize]
  \tikzstyle{every path}=[align = center, font=\scriptsize]

  \node[font = \large\bf] at (4.25, 2.2) {D};

  \node[state] (Imp)    at(0, 2)     {Imp    \\ 3.23};
  \node[state] (ImpArc) at(2.5, 2)   {ImpArc \\ 3.51};
  \node[state] (Ico)    at(4, 0)   {Ico    \\ 3.83};
  \node[state] (Arc)    at(4, 1)     {Arc    \\ 3.80};
  \node[state] (IcoCor) at(0, 1)   {IcoCor \\ 3.23};
  \node[state] (Stan)   at(0, 0)     {Stan   \\ 3.40};

  \path (Imp)    edge [densely dotted] (Arc)
                 edge [densely dotted] (Ico)
        (IcoCor) edge [densely dotted] (Arc)
                 edge [densely dotted] (Ico)
        (Stan)   edge [loosely dotted] (Arc)
                 edge [loosely dotted] (Ico);
\end{tikzpicture}
  \end{minipage}%
  }

  \vspace{0.2cm}

   \fbox{
   \begin{minipage}{.46\linewidth}
\begin{tikzpicture}[->,>=stealth',shorten >=1pt,auto,node distance=2.8cm,
                    semithick]
  \tikzstyle{every state}=[rectangle, align = center, font=\scriptsize]
  \tikzstyle{every path}=[align = center, font=\scriptsize]

  \node[font = \large\bf] at (-0.25, 2.2) {E};

  \node[state] (Imp)    at(3.25, 1) {Imp    \\ 4.33};
  \node[state] (ImpArc) at(1.75, 1)   {ImpArc \\ 3.05};
  \node[state] (Ico)    at(1.75, 2)     {Ico    \\ 2.90};
  \node[state] (Arc)    at(1.75, 0)     {Arc    \\ 3.28};
  \node[state] (IcoCor) at(0, 1)   {IcoCor \\ 2.30};
  \node[state] (Stan)   at(4.5, 1)   {Stan   \\ 5.15};

  \path (IcoCor) edge [dashed]        (Ico)
                 edge                 (ImpArc)
                 edge                 (Arc)
        (Ico)    edge                 (Imp)
                 edge [bend right=80,
                       densely dotted](Arc)
        (ImpArc) edge                 (Imp)
        (Arc)    edge                 (Imp)
        (Imp)    edge                 (Stan);
\end{tikzpicture}
  \end{minipage}%
  }
  \hfill%
  \begin{minipage}{.46\linewidth}
\begin{tikzpicture}[->,>=stealth',shorten >=1pt,auto,node distance=2.8cm,
                    semithick]
  \tikzstyle{every state}=[draw=none,fill=none, align = center]
  \tikzstyle{every path}=[align = center]

  \node[draw=none,fill=none, font = \large] at (0.5, 2.75) {Legend};

  \node[] at (0, 2) (d01) {};
  \node[] at (0, 1.5) (d11) {};
  \node[] at (0, 1) (d21) {};
  \node[] at (0, 0.5) (d31) {};
  \node[right = 1cm of d01] (d02) {};
  \node[right = 1cm of d11] (d12) {};
  \node[right = 1cm of d21] (d22) {};
  \node[right = 1cm of d31] (d32) {};
  \node[draw=none,fill=none, right = 0cm of d02] {significant to niveau $10^{-12}$};
  \node[draw=none,fill=none, right = 0cm of d12] {significant to niveau $10^{-6}$};
  \node[draw=none,fill=none, right = 0cm of d22] {significant to niveau $0.01$};
  \node[draw=none,fill=none, right = 0cm of d32] {significant to niveau $0.1$};

  \path (d01) edge                    (d02)
        (d11) edge   [dashed]         (d12)
        (d21) edge   [densely dotted] (d22)
        (d31) edge   [loosely dotted] (d32);
\end{tikzpicture}

  \end{minipage}%

\caption{Ordering of the kernels in the five situations with respect to their mean ranks (printed below each kernel) over all test instances. A path (possibly using multiple edges) between two kernels denotes a significant difference of the post-hoc Nemenyi test the directions of the arrows follows the ordering of the mean ranks.}

\label{fig:graphs_situations}
\end{figure}
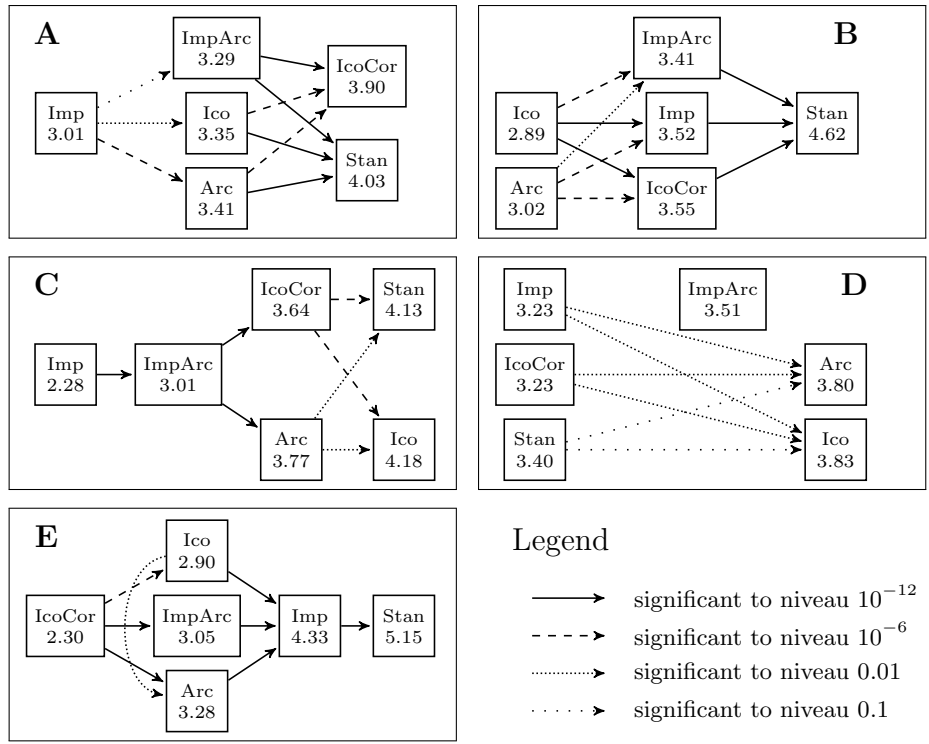

As in the global situation, all Friedman-tests result into very small p-values (numerically approximating zero).
Hence, there is evidence for significant differences between at least some kernels in each situation.
Fig.~\ref{fig:graphs_situations} shows the results of the post-hoc Nemenyi-tests in all five situations.
In situation A and C, the assumption of the Imp-kernel is fulfilled, since $b=0$ allows for imputation.
This is reflected by the results: In both situations A and C the Imp-kernel performs best.
The Imp-kernel outperforms the Arc- and Ico-kernel with a large margin in situation C.
Contrarily, in situation B (unimodal, not imputable) and E (bimodal, not imputable), 
where the imputation assumption is violated, the Imp-kernel performance is inferior. 
These observations fit to our expectations: $b$ controls whether or not
the Imp-kernel is able to find a good value to impute.

The Arc- and the Ico-kernel have similar results in most situations, except for situation C.
This confirms that these kernels encode similar information, and it also shows that the indefiniteness
of the Ico-kernel does not seem to impact optimization performance. At least, the indefiniteness is sufficiently well
mitigated by the employed nugget effect. 

While performing reasonably well, the ImpArc-kernel never achieves a top performance. It is usually positioned in the second-best group.
This can be explained by the fact that it attains some middle ground between the kernels that it combines.
The IcoCorrected-kernel performs poorly in some situations (A, B, C), but it performs best in situation E.
Poor performance may be caused by inconsistencies in the employed definiteness repair methods.
However, it remains unclear to us why the performance is distinctively better in situation E.

Situation D (bimodal, not imputable) has a rather special behavior. Only few distinct differences between the kernels can be detected.
Moreover, it is the only situation in which the Stan kernel does not perform in the worst group.
We suggest that this is due to the fact that modeling the discontinuity is not as important here.
The optimum lies in the region where $x_2$ is active, hence it may even be detrimental to model the discontinuity.
That means, if the optimum is far enough from the discontinuity, it may be helpful to smoothen through the local optimum that lies at the discontinuity, since this will drive the search towards the global optimum.
This could also explain why the Imp-kernel outperforms the Arc-kernel in situation D, despite $b=0.1$.

\section{Conclusion and Outlook}\label{sec:out}
We investigated different kernels for SMBO in hierarchical search spaces, e.g., the Arc-kernel previously proposed by Hutter and Osborne~\cite{Hutter2013}, the Ico-kernel which is similar, yet indefinite, and the Imp-kernel which attempts to learn suitable imputed values for inactive variables.
We tested both the model quality and the optimization performance of six kernels, and received consistent results.
Hence, we can answer our research questions and deduct simple recommendations for choosing a kernel.
\begin{compactenum}
  \item
  The hierarchical structure should be incorporated into the kernel.
  \item
The Imp-kernel should be chosen if it is a-priori known that its assumption is fulfilled.
If the assumption is violated, the Arc- and Ico-kernel are good choices.
Without prior knowledge, the ImpArc-kernel is a sound compromise.
\item
We did not observe many significant differences between the Arc- and the Ico-kernel. 
The kernels' definiteness does not seem to have a strong impact.
\end{compactenum}



These result rely on tests with a rather simple test function, and hence have to interpreted with care.
Devising more complex test functions with higher input dimensions is clearly of interest.
But while artificial tests are instructive due to their controlled behavior, 
it is not always clear how this translates to real world problems.
Hence, it would be desirable to make tests with real world applications,
such as algorithm tuning.

Furthermore, it would be interesting to let the infill optimizer exploit the
information on variable activity,
to avoid searching in inactive areas of the search space.
The same is true for the initialization of the SMBO algorithm. 
Spreading a space-filling design in inactive areas is wasteful.

Finally, all discussed distances $\fdist_i(x_i,x_i')$
are defined for a single dimension $i$. Therefore,
we are not limited to a single choice.
Rather, different distances can be chosen
for each dimension (e.g., Arc for $x_i$, and Imp for $x_{j\neq i}$).

\bibliographystyle{splncs}
\bibliography{article}

\end{document}